\documentclass{article}

\usepackage[preprint,nonatbib]{nips_2018}

\usepackage[utf8]{inputenc} 
\usepackage[T1]{fontenc}    
\usepackage{hyperref}       
\usepackage{url}            
\usepackage{booktabs}       
\usepackage{amsfonts}       
\usepackage{nicefrac}       
\usepackage{microtype}      

\usepackage{bm}
\usepackage{multirow}
\usepackage{amsmath}
\usepackage{amssymb}
\usepackage{amsthm}
\usepackage{color}
\usepackage{times}
\usepackage{graphicx} 
\usepackage{subfigure} 
\usepackage[rightcaption]{sidecap}

\usepackage{algorithm}
\usepackage{algorithmic}

\usepackage{listings}
\definecolor{dkgreen}{rgb}{0,0.6,0}
\definecolor{gray}{rgb}{0.5,0.5,0.5}
\definecolor{mauve}{rgb}{0.58,0,0.82}

\title{PoPPy: A Point Process Toolbox Based on PyTorch}

\author{
  Hongteng Xu\\
  Infinia ML, Inc. and Duke University\\
  \texttt{hongtengxu313@gmail.com} \\
  \includegraphics[width=0.2\linewidth]{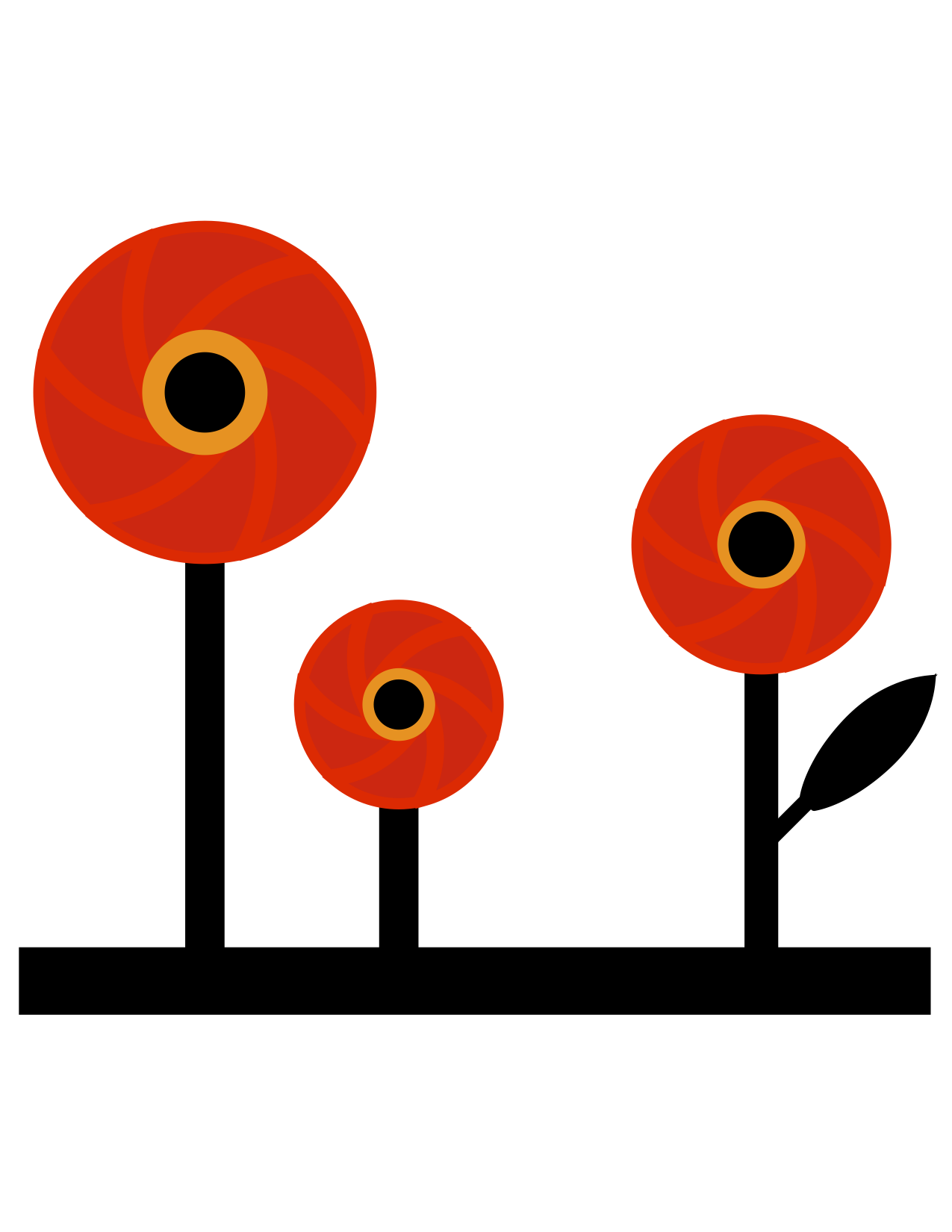}\\
}

\begin{document}

\maketitle

\section{Overview}
\subsection{What is PoPPy?}
\textbf{PoPPy} is a \textbf{Po}int \textbf{P}rocess toolbox based on \textbf{Py}Torch, which achieves flexible designing and efficient learning of point process models. 
It can be used for interpretable sequential data modeling and analysis, $e.g.$, Granger causality analysis of multivariate point processes, point process-based modeling of event sequences, and event prediction. 

\subsection{The Goal of PoPPy}
Many real-world sequential data are often generated by complicated interactive mechanisms among multiple entities. 
Treating the entities as events with different discrete categories, we can represent their sequential behaviors as event sequences in continuous time. 
Mathematically, an event sequence $s$ can be denoted as $\{(t_i^s, c_i^s, f_{c_i^s})\}_{i=1}^{I_s}$, where $t_i^s$ and $c_i^s$ are the timestamp and the event type ($i.e.$, the index of entity) of the $i$-th event, respectively. 
Optionally, each event type may be associated with a feature vector $\bm{f}_c\in \mathbb{R}^{D_c}$, $c\in\mathcal{C}$, and each event sequence may also have a feature vector $\bm{f}_s\in \mathbb{R}^{D_s}$, $s\in\mathcal{S}$. 
Many real-world scenarios can be formulated as event sequences, as shown in Table~\ref{tab:cases}.

\begin{table}[h]
\caption{Typical event sequences in practice.}\label{tab:cases}
\footnotesize{
    \centering
    \begin{tabular}{
        @{\hspace{2pt}}c@{\hspace{2pt}}|
        @{\hspace{2pt}}c@{\hspace{2pt}}|
        @{\hspace{2pt}}c@{\hspace{2pt}}|
        @{\hspace{2pt}}c@{\hspace{2pt}}
        }
        \hline\hline
        Scene &Patient admission &Job hopping &Online shopping\\ \hline
        Entities (Event types) &Diseases &Companies &Items\\
        Sequences &Patients' admission records &LinkedIn users' job history &Buying/rating behaviors\\
        Event feature &Diagnose records &Job descriptions &Item profiles\\
        Sequence feature &Patient profiles &User profiles &User profiles\\
        Task &Build Disease network &Model talent flow &Recommendation system\\
        \hline\hline
      \end{tabular}
}
\end{table}

Given a set of event sequences $\mathcal{S}$, we aim to model the dynamics of the event sequences, capture the interactive mechanisms among different entities, and predict their future behaviors. 
Temporal point processes provide us with a potential solution to achieve these aims. 
In particular, a multivariate temporal point process can be represented by a set of counting processes $N=\{N_c(t)\}_{c\in\mathcal{C}}$, in which $N_c(t)$ is the number of type-$c$ events occurring till time $t$. 
For each $N_c(t)$, the expected instantaneous happening rate of type-$c$ events at time $t$ is denoted as $\lambda_c(t)$, which is called ``\textbf{intensity function}'':
\begin{eqnarray}
\begin{aligned}
\lambda_c(t) = \frac{\mathbb{E}[dN_c(t)|\mathcal{H}_t]}{dt},~\mathcal{H}_t= \{(t_i, c_i) |  t_i < t, c_i\in\mathcal{C}\},
\end{aligned}
\end{eqnarray}
where $\mathcal{H}_t$ represents historical observations before time $t$. 

\begin{figure}[h]
\centering
\includegraphics[width=0.8\textwidth]{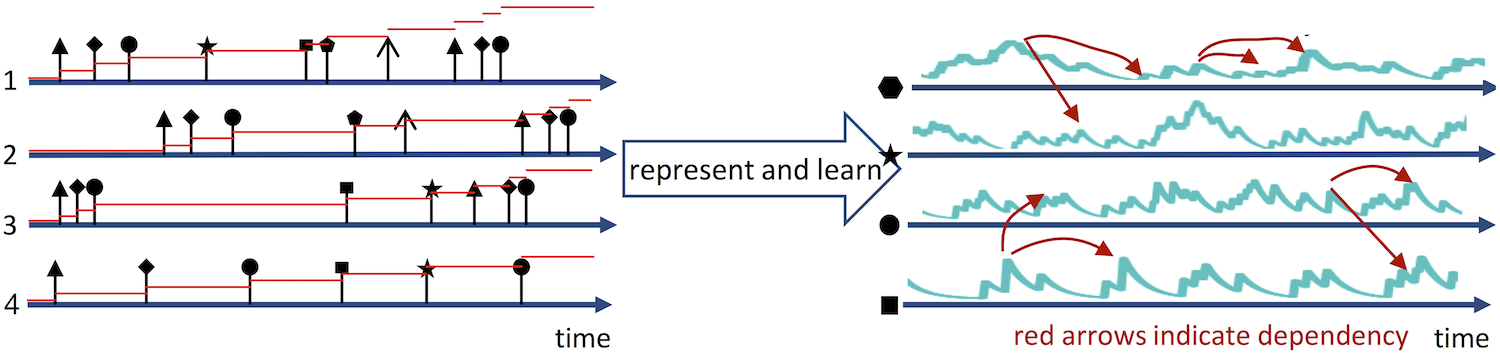}
\caption{Event sequences and intensity functions.}\label{fig:illu}
\end{figure}

As shown in Fig.~\ref{fig:illu}, the counting processes can be represented as a set of intensity functions, each of which corresponds to a specific event type. 
The temporal dependency within the same event type and that across different event types ($i.e.$, the red arrows in Fig.~\ref{fig:illu}) can be captured by choosing particular intensity functions. 
Therefore, the key points of point process-based sequential data modeling include
\begin{enumerate}
    \item How to design intensity functions to describe the mechanism behind observed data?
    \item How to learn the proposed intensity functions from observed data?
\end{enumerate}

\textbf{The goal of PoPPy is providing a user-friendly solution to the key points above and achieving large-scale point process-based sequential data analysis, simulation, and prediction. }

\subsection{Installation of PoPPy}
\begin{figure}[h]
\centering
\includegraphics[width=0.7\textwidth]{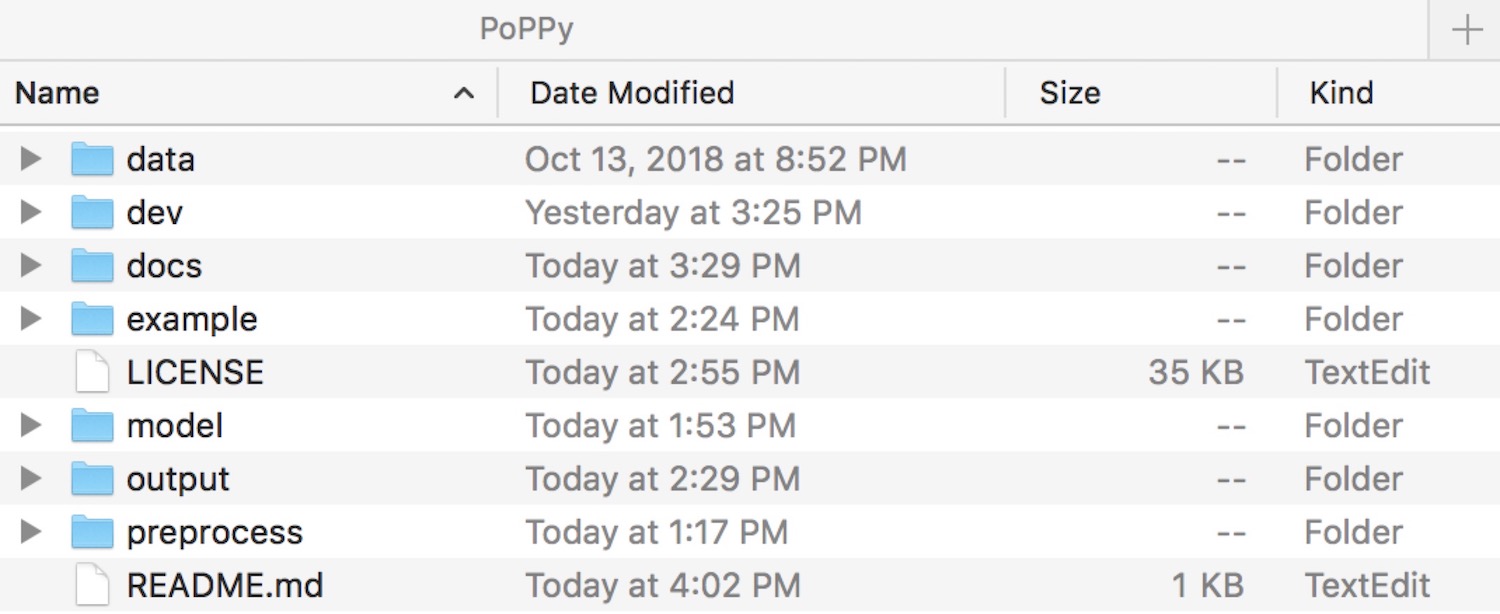}
\caption{The subfolders in the package of PoPPy.}\label{fig:folder}
\end{figure}

\begin{figure}[h]
\centering
\includegraphics[width=0.7\textwidth]{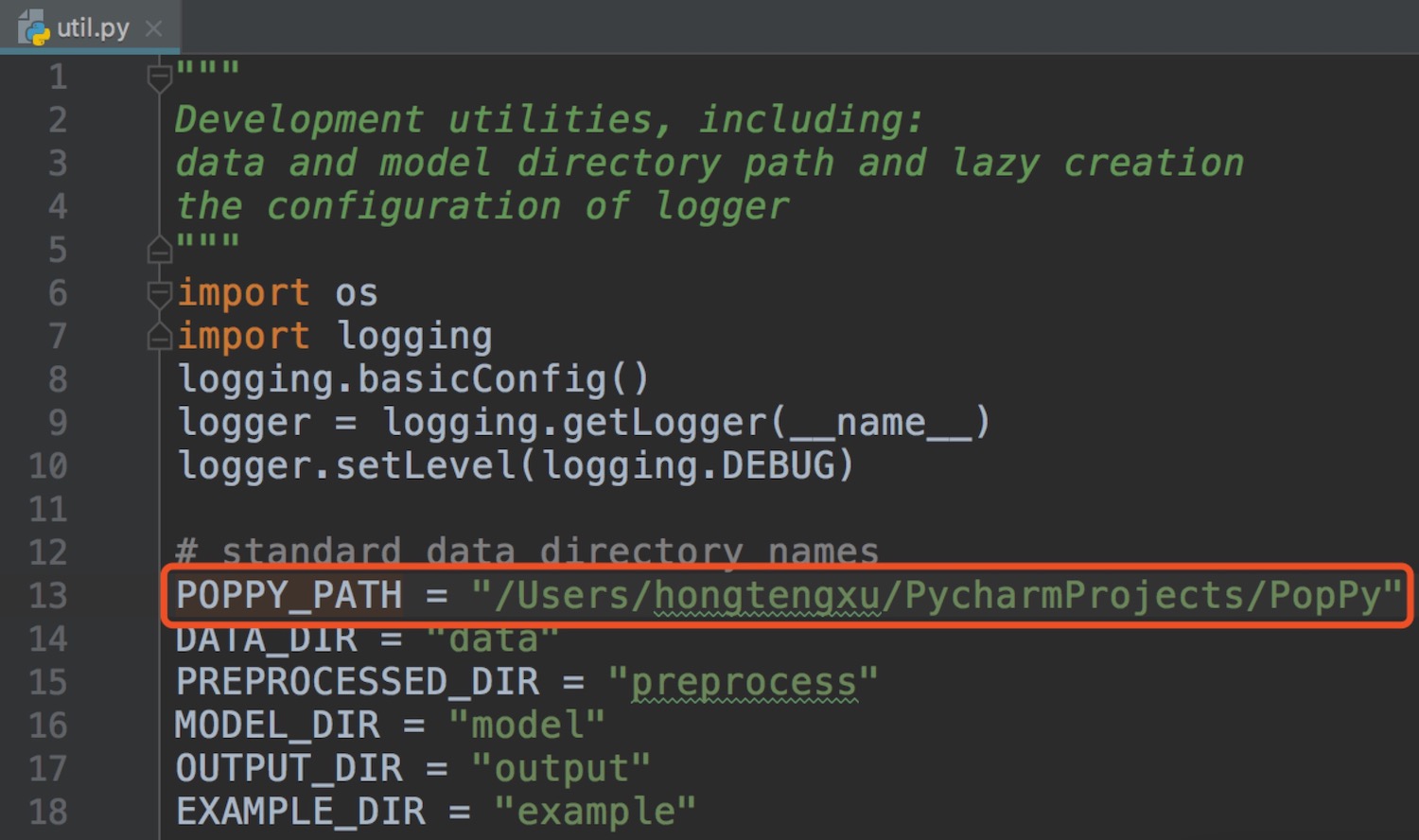}
\caption{An example of the path of PoPPy.}\label{fig:path}
\end{figure}

PoPPy is developed on Mac OS 10.13.6 but also tested on Ubuntu 16.04.  
The installation of PoPPy is straightforward. In particular,
\begin{enumerate}
    \item Install Anaconda3 and create a conda environment.
    \item Install PyTorch0.4 in the environment.
    \item Download PoPPy from \url{https://github.com/HongtengXu/PoPPy/} and unzip it to the directory in the environment. 
    The unzipped folder should contain several subfolders, as shown in Fig.~\ref{fig:folder}.
    \item Open \texttt{dev/util.py} and change \textsf{POPPY\_PATH} to the directory, as shown in Fig.~\ref{fig:path}.
\end{enumerate}

The subfolders in the package include
\begin{itemize}
    \item \texttt{data:} It contains a toy dataset in \emph{.csv} format.
    \item \texttt{dev:} It contains a \texttt{util.py} file, which configures the path and the logger of the package.
    \item \texttt{docs:} It contains the tutorial of PoPPy.
    \item \texttt{example:} It contains some demo scripts for testing the functionality of the package.
    \item \texttt{model:} It contains the classes of predefined point process models and their modules.
    \item \texttt{output:} It contains the output files generated by the demo scripts in the  \texttt{example} folder.
    \item \texttt{preprocess:} It contains the classes and the functions of data I/O and preprocessing.
\end{itemize}
In the following sections, we will introduce the details of PoPPy.

\section{Data: Representation and Preprocessing}
\subsection{Representations of Event Sequences}
PoPPy represents observed event sequences as a nested dictionary. 
In particular, the proposed \texttt{database} has the following structure:
\begin{lstlisting}
database = {
    'event_features'    : None or (De, C) float array of event features,
                            C is the number of event types.
                            De is the dimension of event feature.
    'type2idx'          : a Dict = {'event_name': event_index}
    'idx2type'          : a Dict = {event_index: 'event_name'}
    'seq2idx'           : a Dict = {'seq_name': seq_index}
    'idx2seq'           : a Dict = {seq_index: 'seq_name'}
    'sequences'         : a List = [seq_1, seq_2, ..., seq_N].
}

For the i-th sequence:
seq_i = {
    'times'             : (N,) float array of timestamps, 
                            N is the number of events.
    'events'            : (N,) int array of event types.
    'seq_feature'       : None or (Ds,) float array of sequence feature.
                            Ds is the dimension of sequence feature
    't_start'           : a float number, the start timestamp of the sequence.
    't_stop'            : a float number, the stop timestamp of the sequence.
    'label'             : None or int/float number, the labels of the sequence
}
\end{lstlisting}

PoPPy provides three functions to load data from \emph{.csv} file and convert it to the proposed \texttt{database}.

\subsubsection{\texttt{preprocess.DataIO.load\_sequences\_csv}}
\begin{figure}[h]
    \centering
    \includegraphics[width=0.9\linewidth]{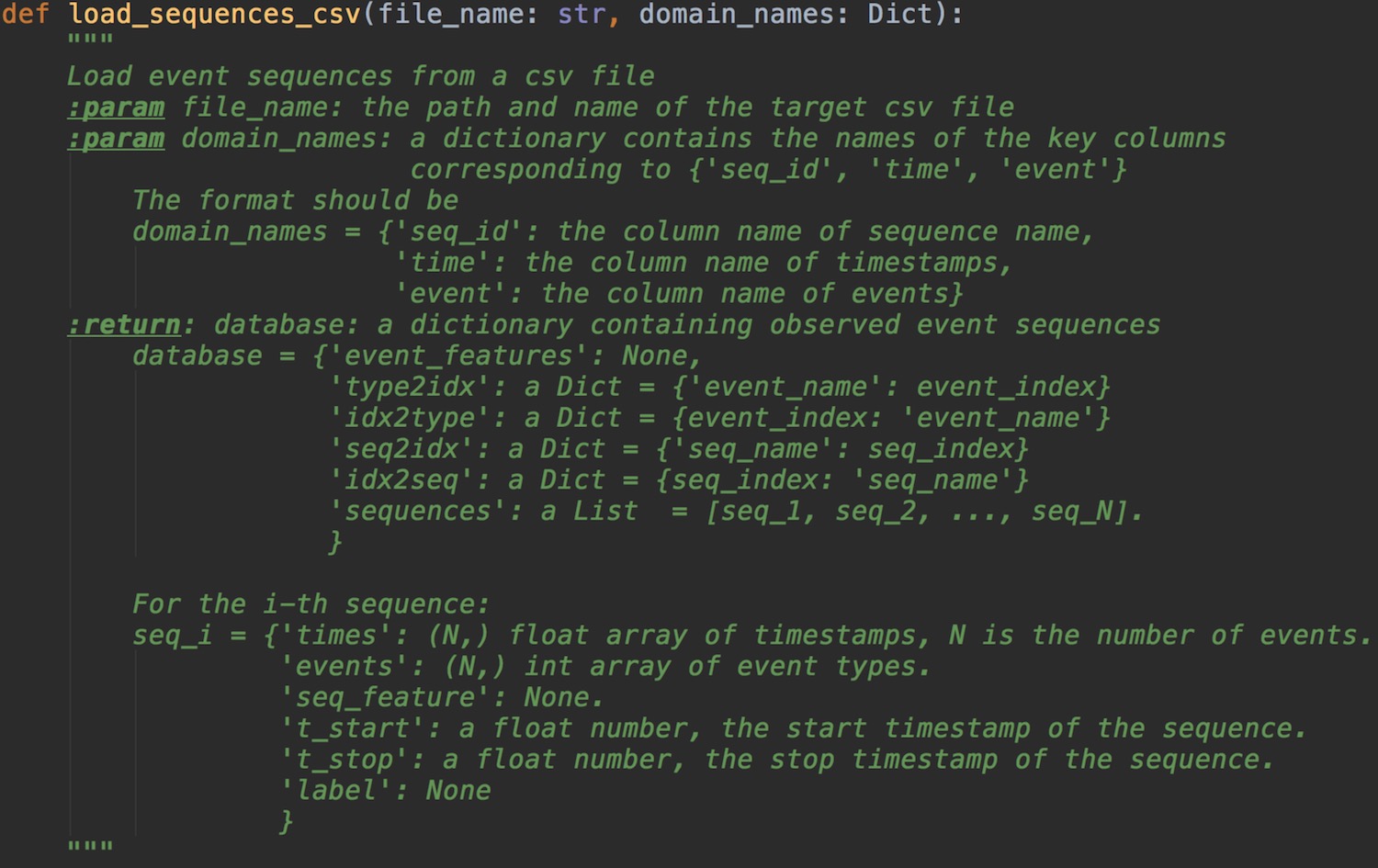}
    \caption{The description of \texttt{load\_sequences\_csv}.}
    \label{fig:loadseq}
\end{figure}

This function loads event sequences and converts them to the proposed \texttt{database}. 
The IO and the description of this function are shown in Fig.~\ref{fig:loadseq}.

\begin{figure}[h]
    \centering
    \includegraphics[width=0.6\linewidth]{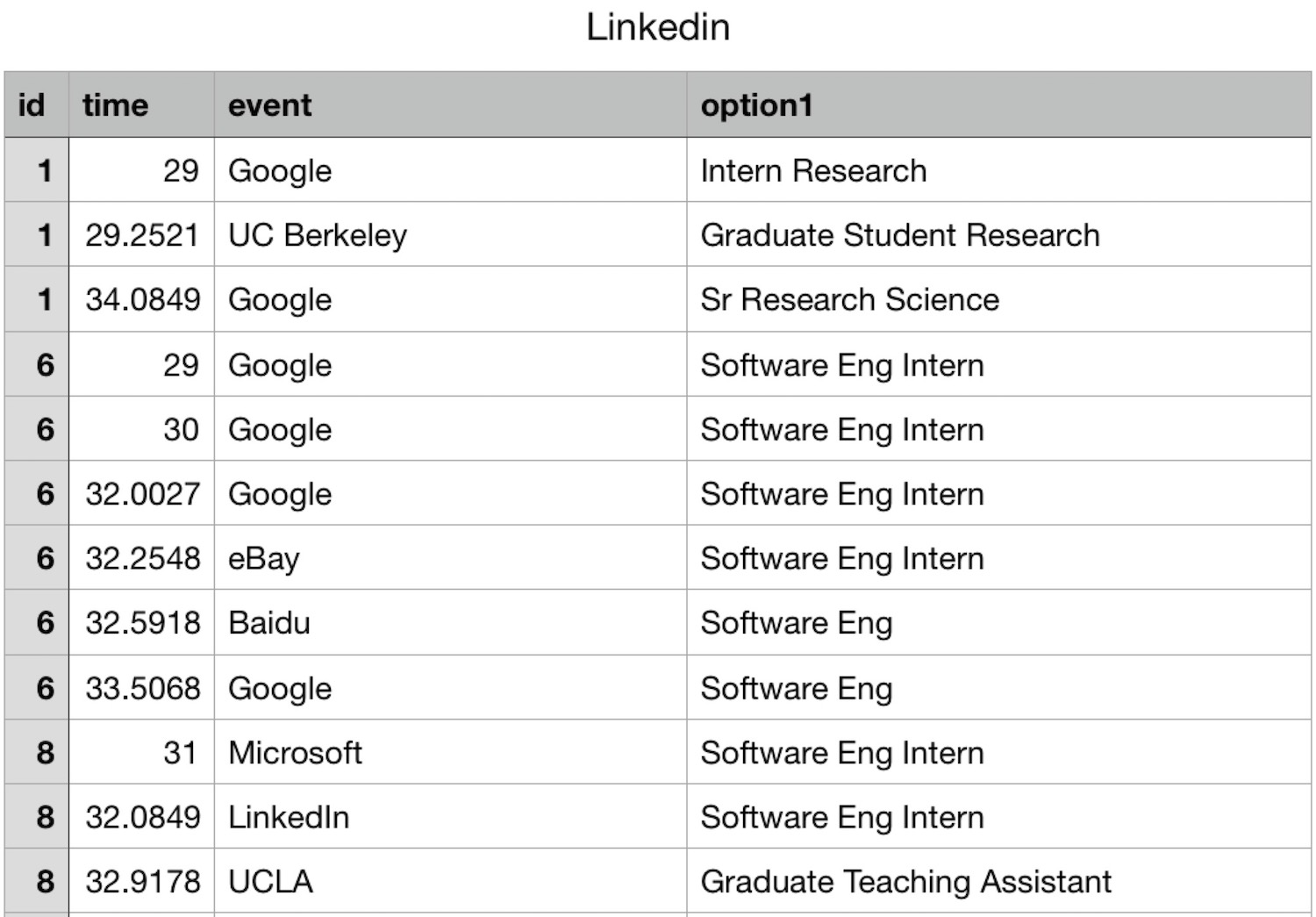}
    \caption{Some rows of \emph{Linkedin.csv}.}
    \label{fig:linkedin}
\end{figure}

For example, the \emph{Linkedin.csv} file in the folder \texttt{data} records a set of linkedin users' job-hopping behaviors among different companies, whose format is shown in Fig.~\ref{fig:linkedin}.

Here, the column \textsf{id} corresponds to the names of sequences ($i.e.$ the index of users), the column \textsf{time} corresponds to the timestamps of events ($i.e.$ the ages that the users start to work), and the column \textsf{event} corresponds to the event types ($i.e.$, the companies). 
Therefore, we can define the input \texttt{domain\_names} as
\begin{lstlisting}
domain_names = {
    'seq_id'    : 'id',
    'time'      : 'time',
    'event'     : 'event'
}
\end{lstlisting}
and \texttt{database = load\_sequences\_csv('Linkedin.csv', domain\_names)}.

Note that the \texttt{database} created by \texttt{load\_sequences\_csv()} does not contain event features and sequence features, whose values in \texttt{database} are \textbf{None}. 
PoPPy supports to load categorical or numerical features from \emph{.csv} files, as shown below.

\subsubsection{\texttt{preprocess.DataIO.load\_seq\_features\_csv}}
\begin{figure}[h]
    \centering
    \includegraphics[width=1\linewidth]{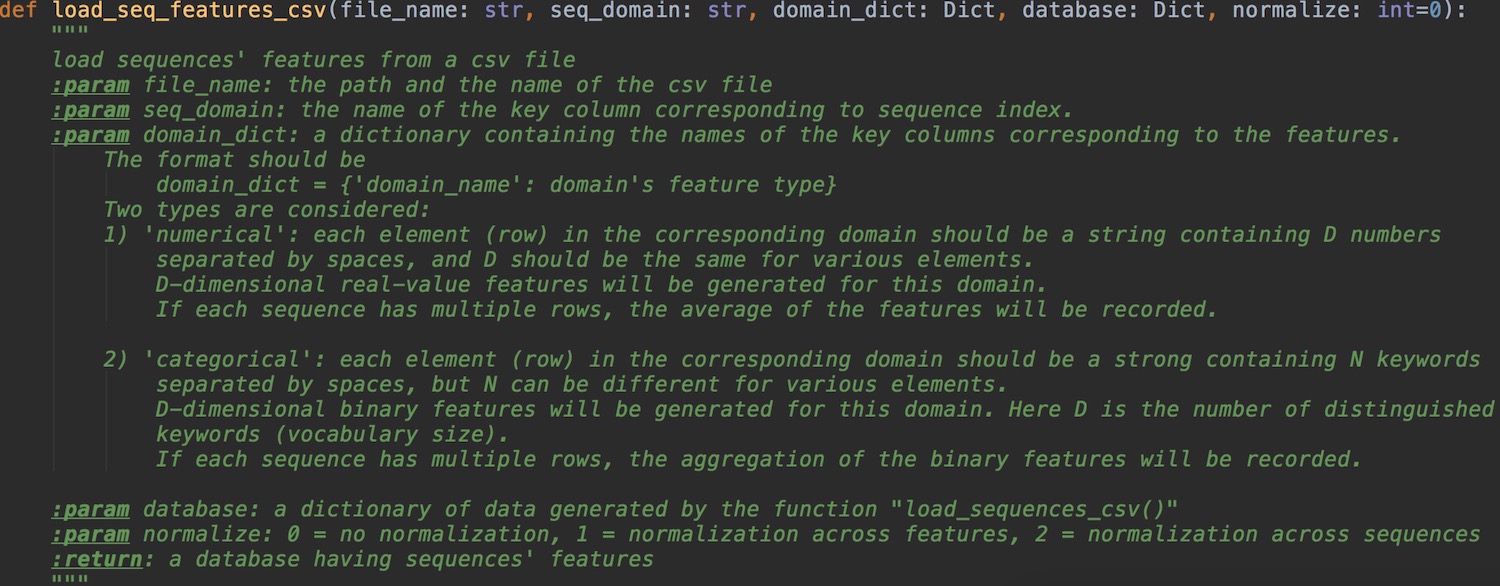}
    \caption{The description of \texttt{load\_seq\_features\_csv}.}
    \label{fig:loadseqfeat}
\end{figure}

This function loads sequence features from a \emph{.csv} file and import them to the proposed \texttt{database}. 
The IO and the description of this function are shown in Fig.~\ref{fig:loadseqfeat}.
Take the \emph{Linkedin.csv} file as an example. 
Suppose that we have already create \texttt{database} by the function \texttt{load\_sequences\_csv}, and we want to take the column \textsf{option1} ($i.e.$, the job titles that each user had) as the categorical features of event sequences. 
We should have
\begin{lstlisting}
domain_names = {
    'option1'    : 'categorical'
}
database = load_seq_features_csv(
                                file_name = 'Linkedin.csv',
                                seq_domain = 'seq_id',
                                domain_dict = domain_names,
                                database = database)
\end{lstlisting}

Here the input \texttt{normalize} is set as default $0$, which means that the features in \texttt{database['sequences'][i]['seq\_feature']}, $i=1,...,|\mathcal{S}|$, are not normalized.

\subsubsection{\texttt{preprocess.DataIO.load\_event\_features\_csv}}
\begin{figure}[h]
    \centering
    \includegraphics[width=1\linewidth]{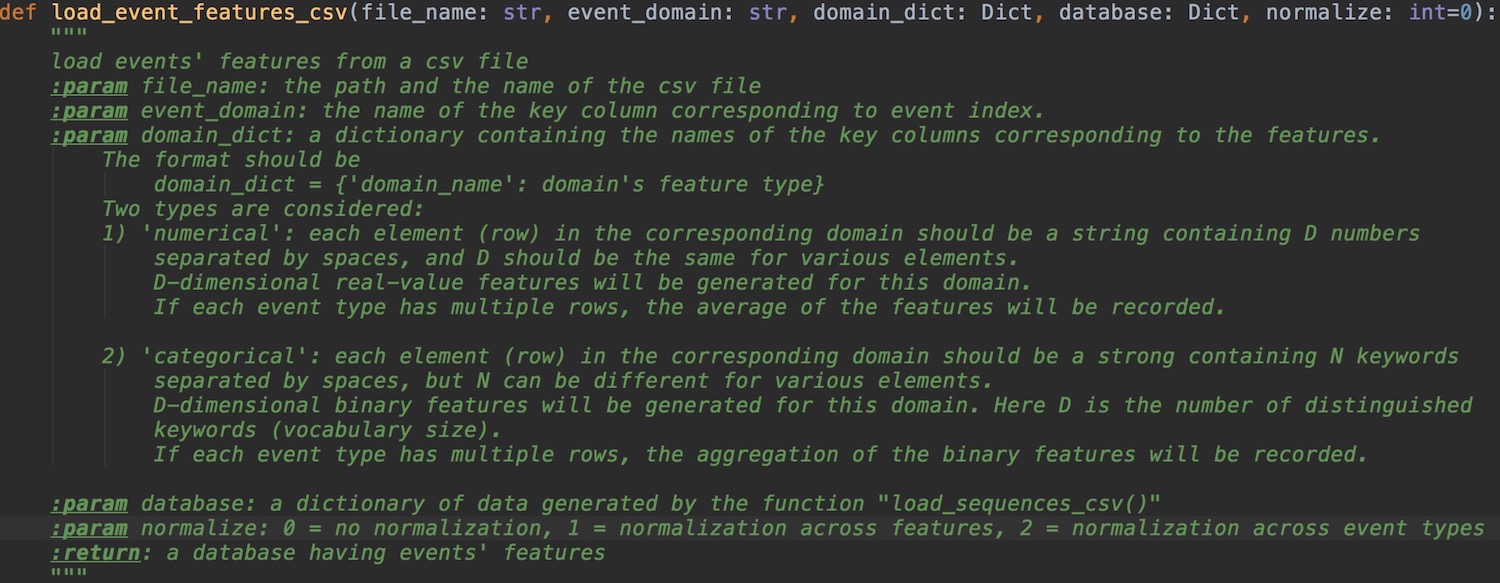}
    \caption{The description of \texttt{load\_event\_features\_csv}.}
    \label{fig:loadeventfeat}
\end{figure}

This function loads event features from a \emph{.csv} file and import them to the proposed \texttt{database}.
The IO and the description of this function are shown in Fig.~\ref{fig:loadeventfeat}.
Similarly, if we want to take the column \textsf{option1} in \emph{Linkedin.csv} as the categorical features of event types, we should have
\begin{lstlisting}
domain_names = {
    'option1'    : 'categorical'
}
database = load_event_features_csv(
                                file_name = 'Linkedin.csv',
                                event_domain = 'event',
                                domain_dict = domain_names,
                                database = database)
\end{lstlisting}

\subsection{Operations for Data Preprocessing}
\begin{figure}[h]
    \centering
    \includegraphics[width=0.80\linewidth]{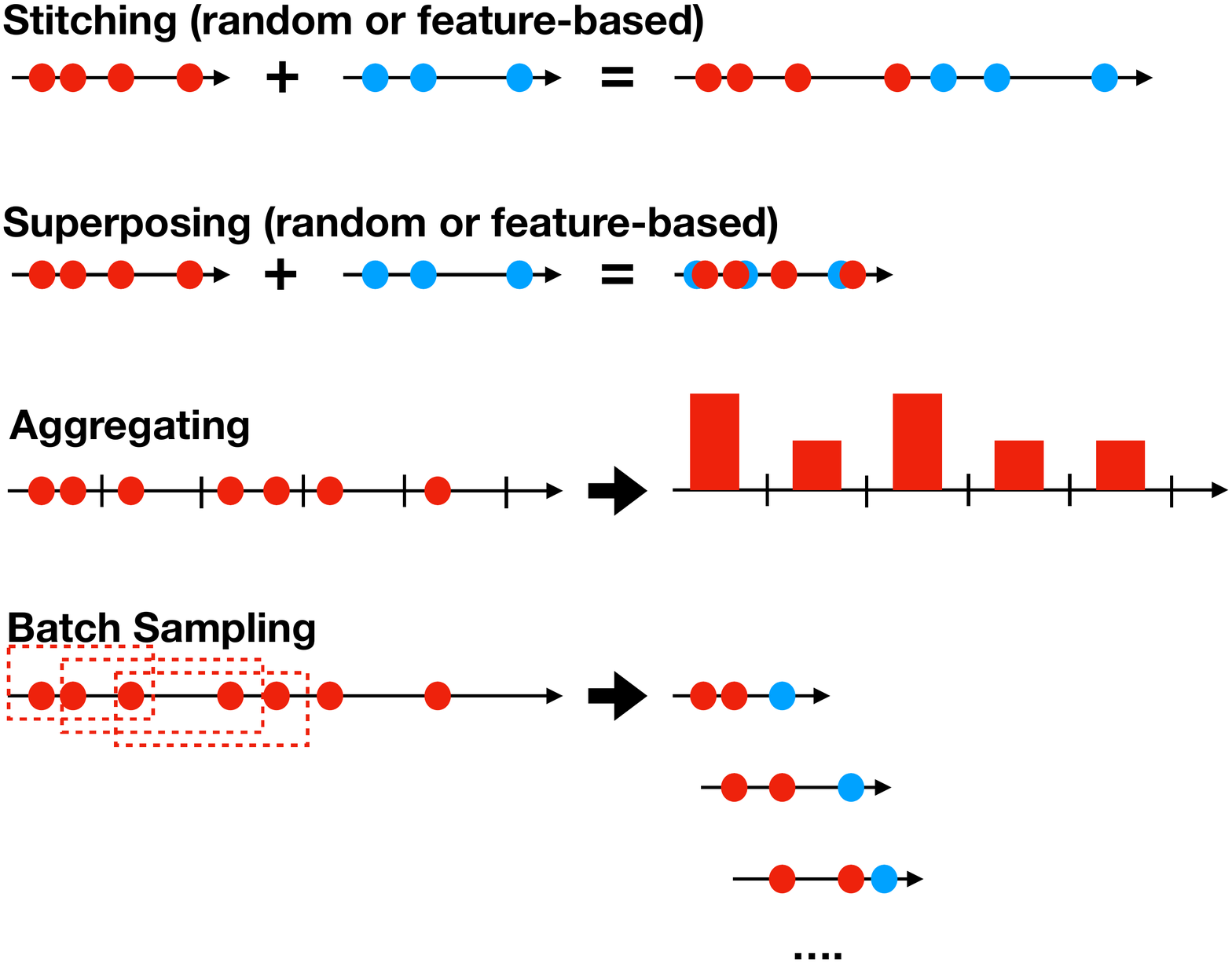}
    \caption{The illustration of four data operations.}
    \label{fig:dataops}
\end{figure}

Besides necessary sequence/feature loaders and converters mentioned above, PoPPy contains multiple useful functions and classes for data preprocessing, including sequence stitching, superposing, aggregating, and batch sampling. 
Fig.~\ref{fig:dataops} illustrates the corresponding data operations. 

\subsubsection{\texttt{preprocess.DataOperation.stitching}}
\begin{figure}[h]
    \centering
    \includegraphics[width=1\linewidth]{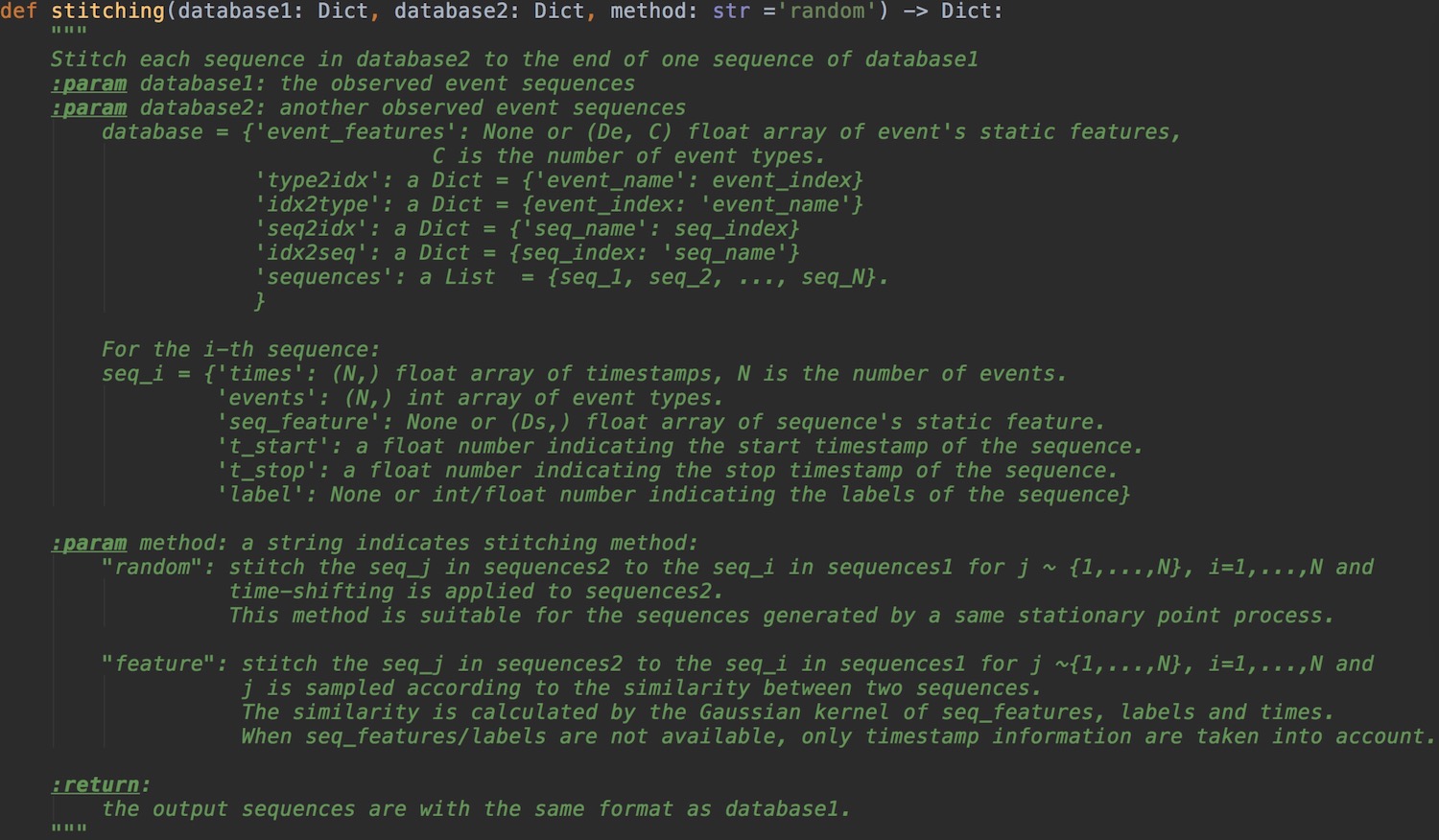}
    \caption{The description of \texttt{stitching}.}
    \label{fig:stitch}
\end{figure}

This function stitches the sequences in two \texttt{database} randomly or based on their \texttt{seq\_feature} and time information (\texttt{t\_start}, \texttt{t\_stop}). 
Its description is shown in Fig.~\ref{fig:stitch}. 

When \texttt{method = 'random'}, for each sequence in \texttt{database1} the function randomly selects a sequence in \texttt{database2} as its follower and stitches them together. 
When \texttt{method = 'feature'}, the similarity between the sequence in \texttt{database1} and that in \texttt{database2} is defined by the multiplication of a temporal Gaussian kernel and a sequence feature's Gaussian kernel, and the function selects the sequence in \texttt{database2} yielding to a distribution defined by the similarity. The stitching method has been proven to be useful for enhancing the robustness of learning results, especially when the training sequences are very short~\cite{xu2017learning,xu2018learning}. 

\subsubsection{\texttt{preprocess.DataOperation.superposing}}
\begin{figure}[h]
    \centering
    \includegraphics[width=1\linewidth]{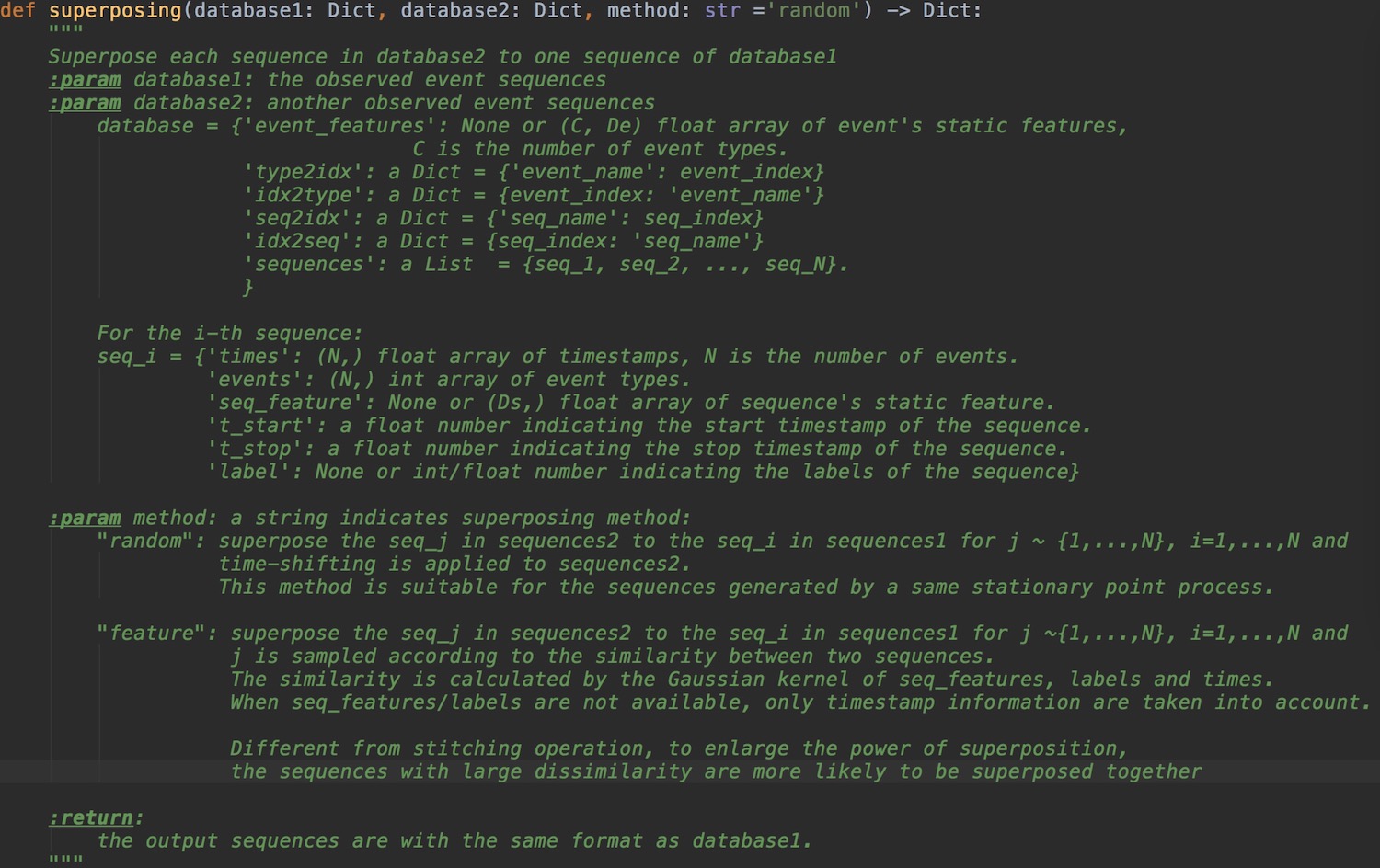}
    \caption{The description of \texttt{superposing}.}
    \label{fig:superpose}
\end{figure}

This function superposes the sequences in two \texttt{database} randomly or based on their \texttt{seq\_feature} and time information (\texttt{t\_start}, \texttt{t\_stop}). 
Its description is shown in Fig.~\ref{fig:superpose}. 

When \texttt{method = 'random'}, for each sequence in \texttt{database1} the function randomly selects a sequence in \texttt{database2} and superposes them together. 
When \texttt{method = 'feature'}, the similarity between the sequence in \texttt{database1} and that in \texttt{database2} is defined by the multiplication of a temporal Gaussian kernel and a sequence feature's Gaussian kernel, and the function selects the sequence in \texttt{database2} yielding to a distribution defined by the similarity. 

Similar to the stitching operation, the superposing method has been proven to be useful for learning linear Hawkes processes robustly. 
However, it should be noted that different from stitching operation, which stitches similar sequences with a high probability, the superposing process would like to superpose the dissimilar sequences with a high chance. 
The rationality of such an operation can be found in my paper~\cite{xu2018benefits,xu2018superposition}. 

\subsubsection{\texttt{preprocess.DataOperation.aggregating}}
\begin{figure}[h]
    \centering
    \includegraphics[width=1\linewidth]{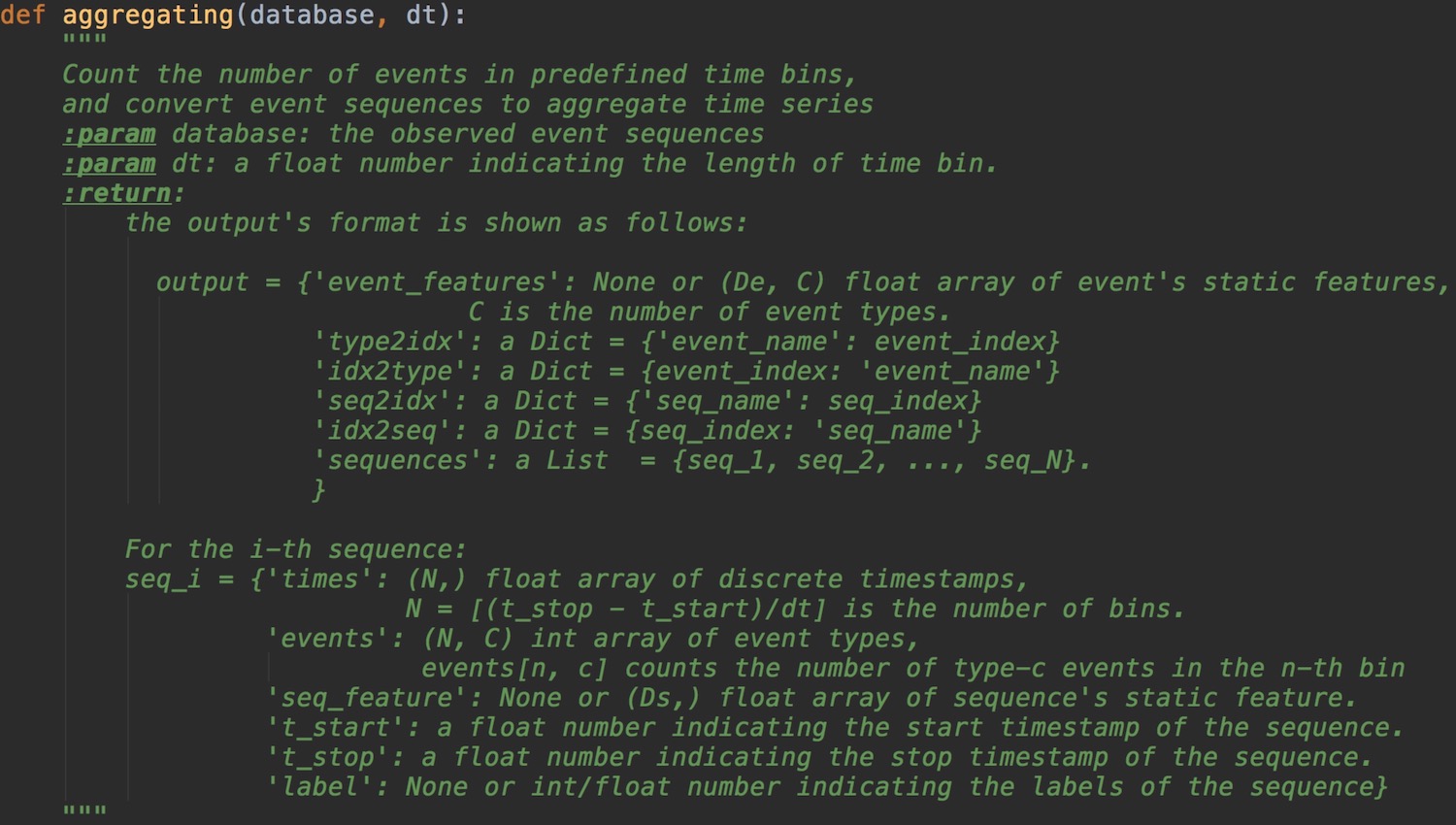}
    \caption{The description of \texttt{aggregate}.}
    \label{fig:aggregate}
\end{figure}

This function discretizes each event sequence into several bins and counts the number of events with specific types in each bin. 
Its description is shown in Fig.~\ref{fig:aggregate}.

\subsubsection{\texttt{preprocess.DataOperation.EventSampler}}
This class is a subclass of \texttt{torch.utils.data.Dataset}, which samples batches from \texttt{database}. 
For each sample in the batch, an event ($i.e.$, its event type and timestamp) and its history with length \texttt{memorysize} ($i.e.$, the last \texttt{memorysize} events and their timestamps) are recorded. 
If the features of events (or sequences) are available, the sample will record these features as well.
\begin{figure}[h]
    \centering
    \includegraphics[width=1\linewidth]{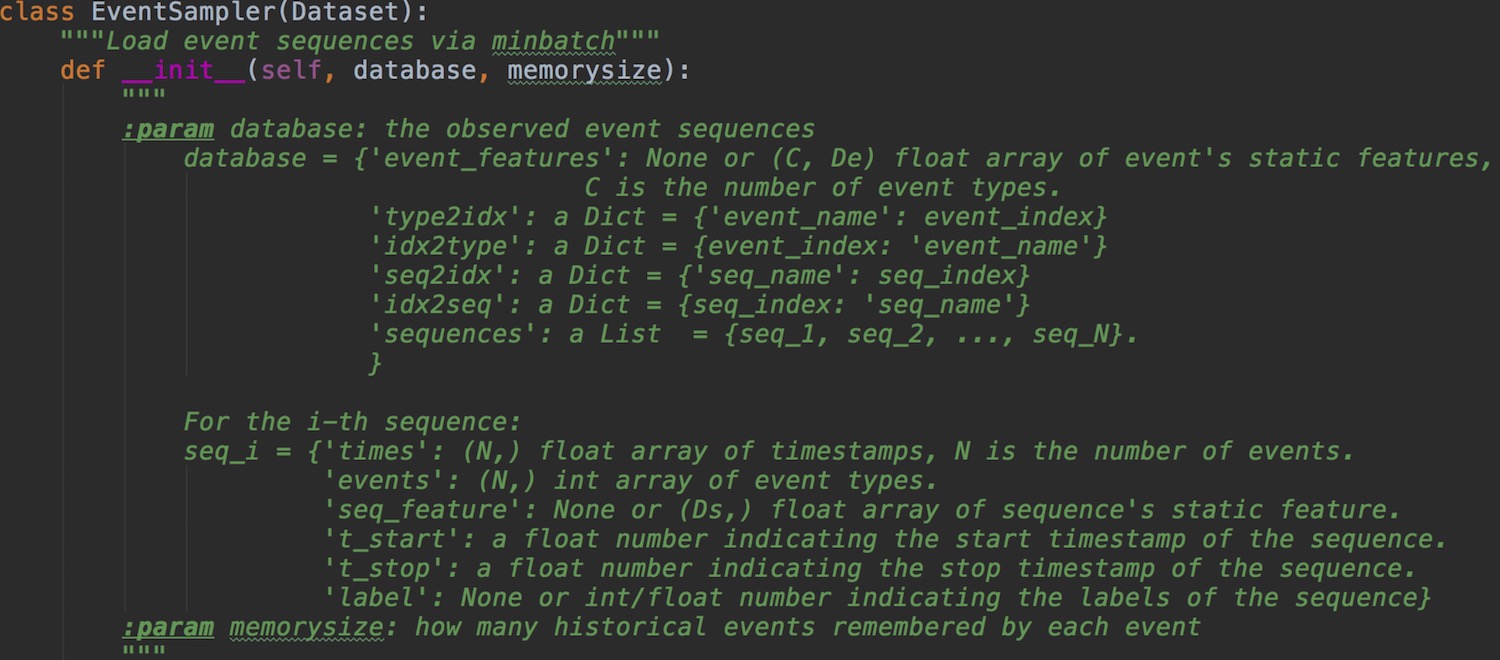}
    \caption{The description of \texttt{EventSampler}.}
    \label{fig:eventsampler}
\end{figure}

\section{Temporal Point Process Models}
\subsection{Modular design of point process model}
\begin{figure}[h]
\centering
\includegraphics[width=1\textwidth]{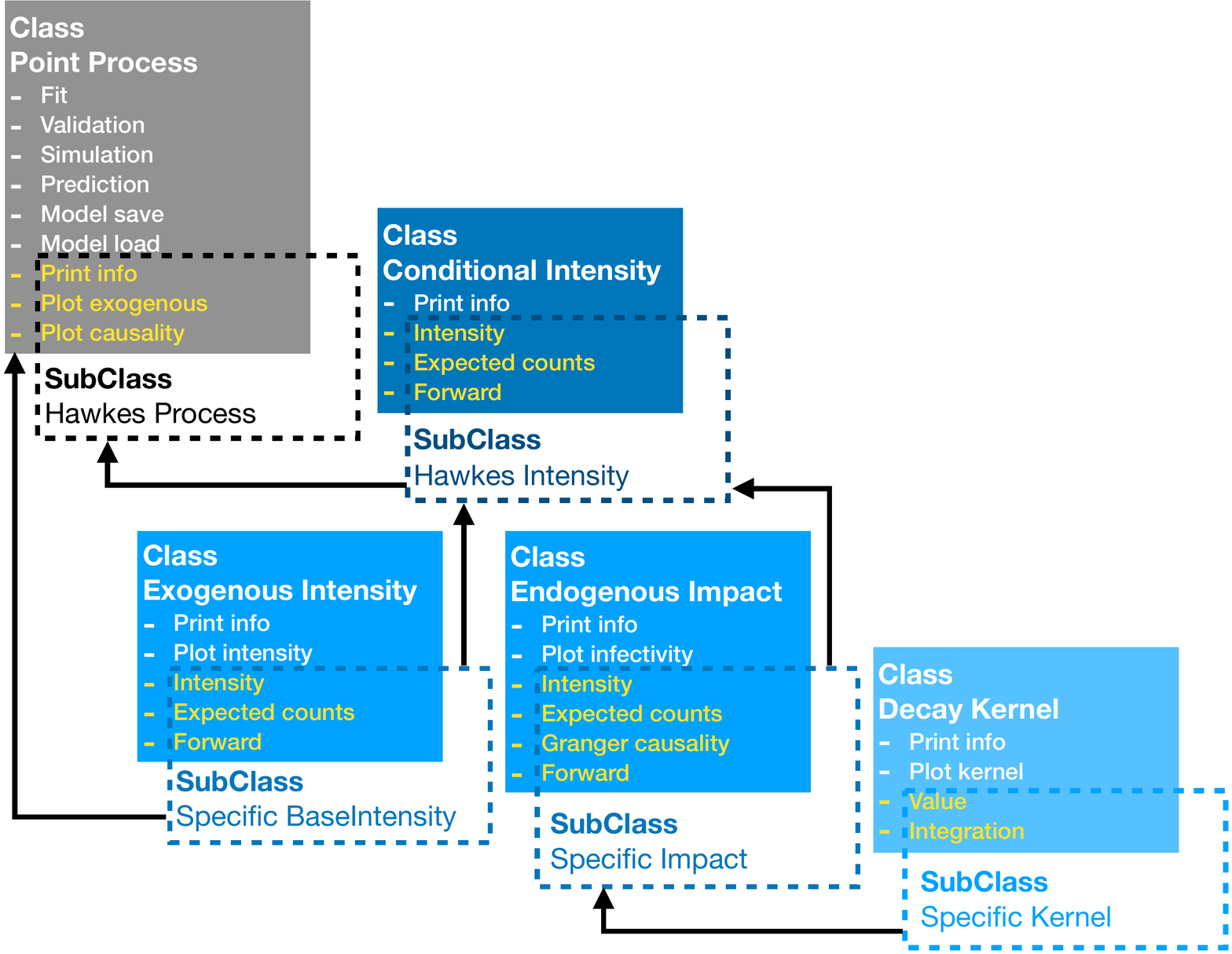}
\caption{An illustration of proposed modular design strategy. Each color block represents a class with some functions. 
For each block, the dotted frame represents one of its subclass, which inherits some functions (the white ones) while overrides some others or creates new ones (the yellow ones).
The black arrow means that the destination class will call the instance of the source class as input.}\label{fig:design}
\end{figure}

PoPPy applies a flexible strategy to build point process's intensity functions from interpretable modules. 
Such a modular design strategy is very suitable for the Hawkes process and its variants. 
Fig.~\ref{fig:design} illustrates the proposed modular design strategy. 
In the following sections, we take the Hawkes process and its variants as examples and introduce the modules ($i.e.$, the classes) in PoPPy. 

\subsection{\texttt{model.PointProcess.PointProcessModel}}
This class contains basic functions of a point process model, including
\begin{itemize}
    \item \texttt{fit:} learn model's parameters given training data. It description is shown in Fig.~\ref{fig:fit}
    \begin{figure}[h]
    \centering
    \includegraphics[width=1\textwidth]{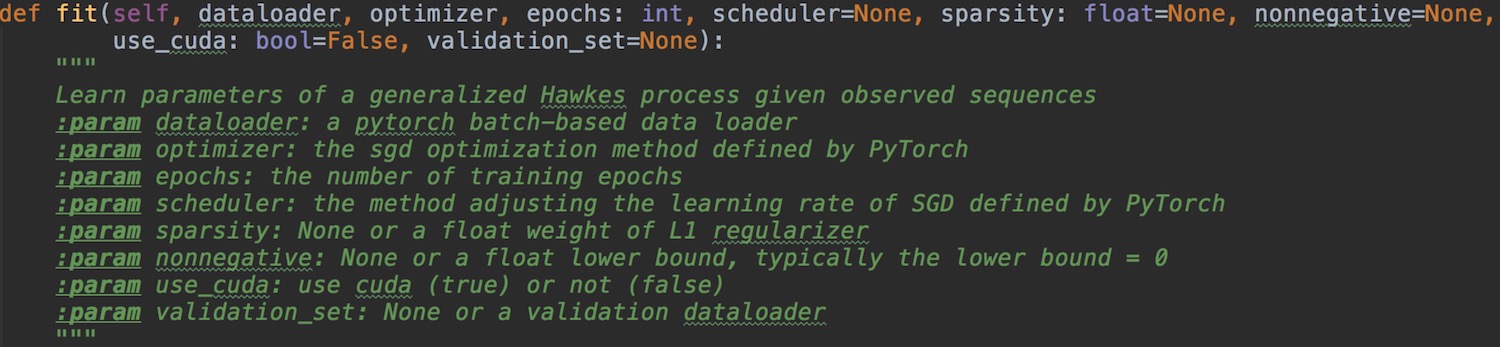}
    \caption{The description of \texttt{fit}.}\label{fig:fit}
    \end{figure}
    \item \texttt{validation:} test model given validation data. It description is shown in Fig.~\ref{fig:valid}
    \begin{figure}[h]
    \centering
    \includegraphics[width=1\textwidth]{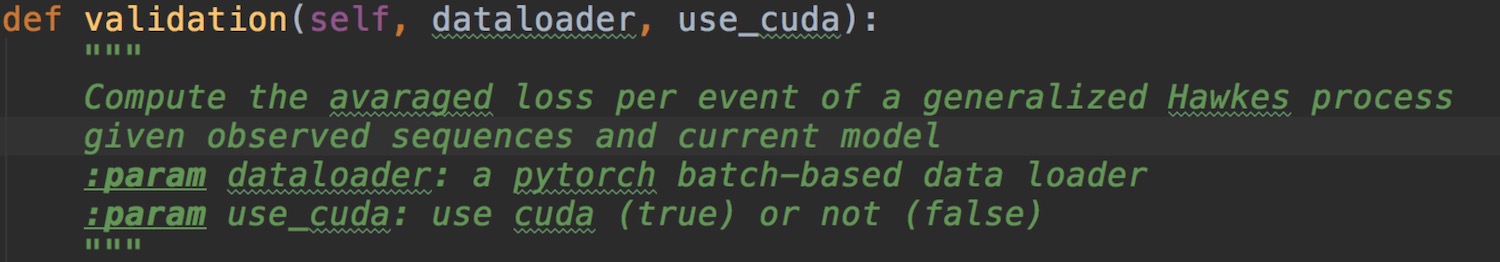}
    \caption{The description of \texttt{validation}.}\label{fig:valid}
    \end{figure}
    \item \texttt{simulation:} simulate new event sequences from scratch or following observed sequences by Ogata's thinning algorithm~\cite{ogata1981lewis}. It description is shown in Fig.~\ref{fig:sim}
    \begin{figure}[h]
    \centering
    \includegraphics[width=1\textwidth]{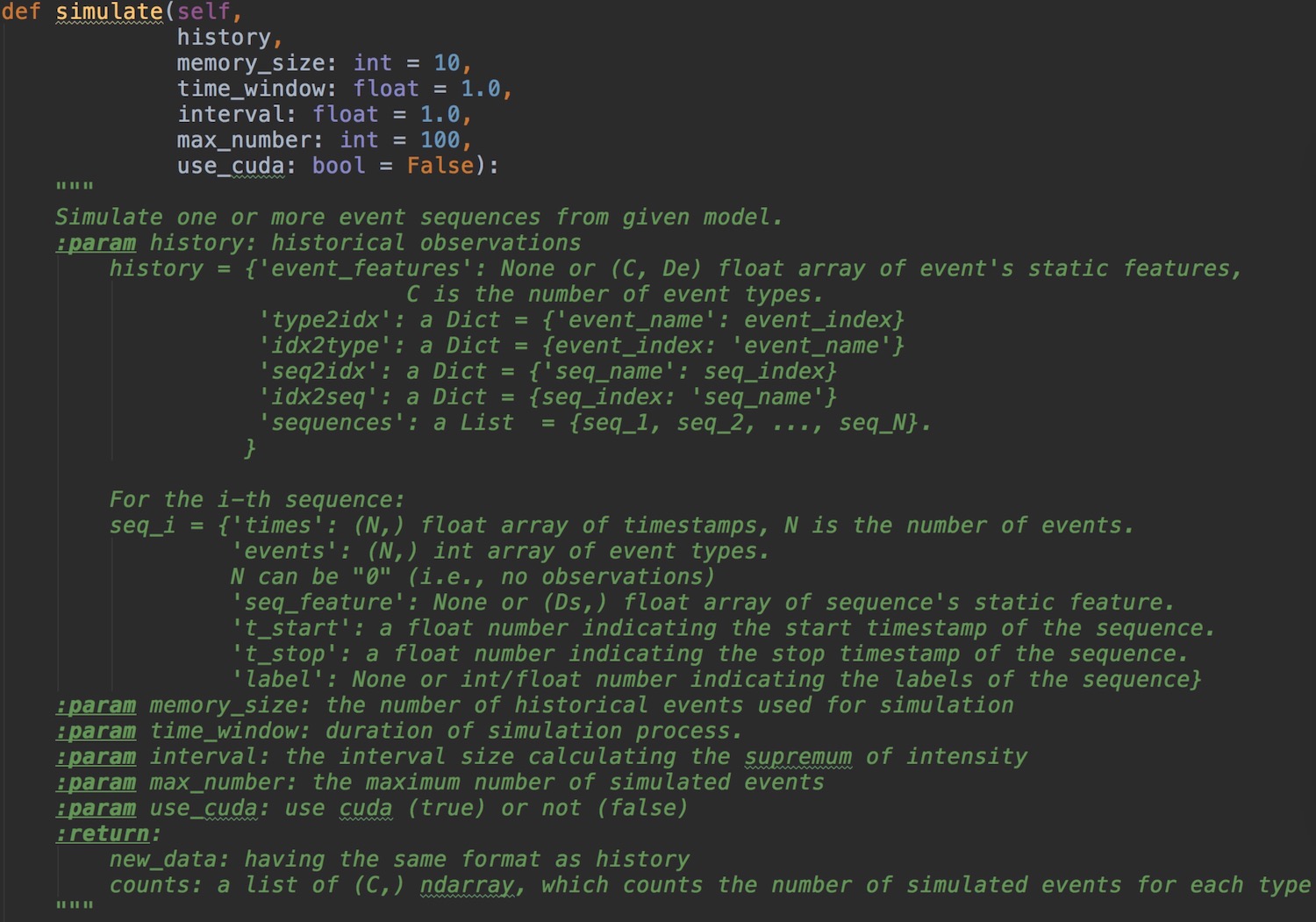}
    \caption{The description of \texttt{simulate}.}\label{fig:sim}
    \end{figure}
    \item \texttt{prediction:} predict expected counts of the events in the target time inteveral given learned model and observed sequences. It description is shown in Fig.~\ref{fig:pred}
    \begin{figure}[h]
    \centering
    \includegraphics[width=1\textwidth]{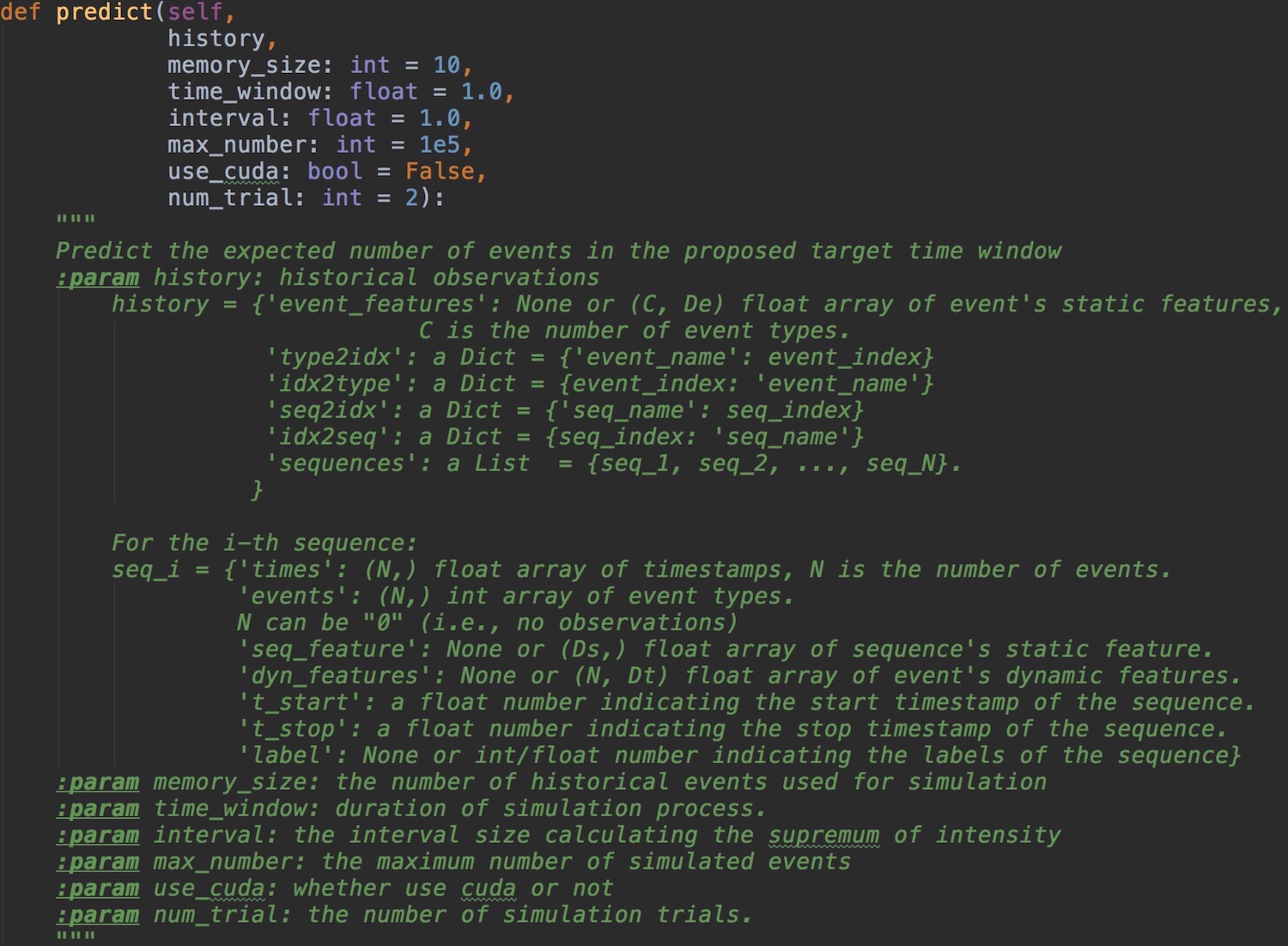}
    \caption{The description of \texttt{predict}.}\label{fig:pred}
    \end{figure}
    \item \texttt{model\_save:} save model or save its parameters.
    It description is shown in Fig.~\ref{fig:save}
    \begin{figure}[h]
    \centering
    \includegraphics[width=1\textwidth]{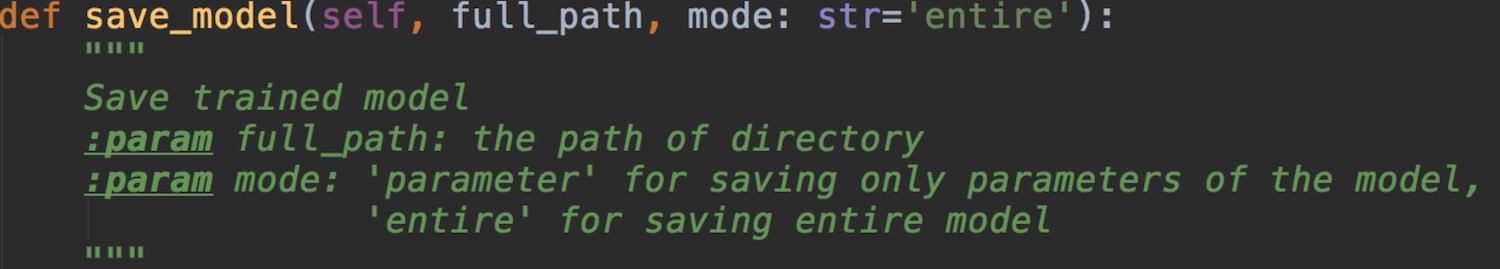}
    \caption{The description of \texttt{model\_save}.}\label{fig:save}
    \end{figure}
    \item \texttt{model\_load:} load model or load its parameters. It description is shown in Fig.~\ref{fig:load}
    \begin{figure}[h]
    \centering
    \includegraphics[width=1\textwidth]{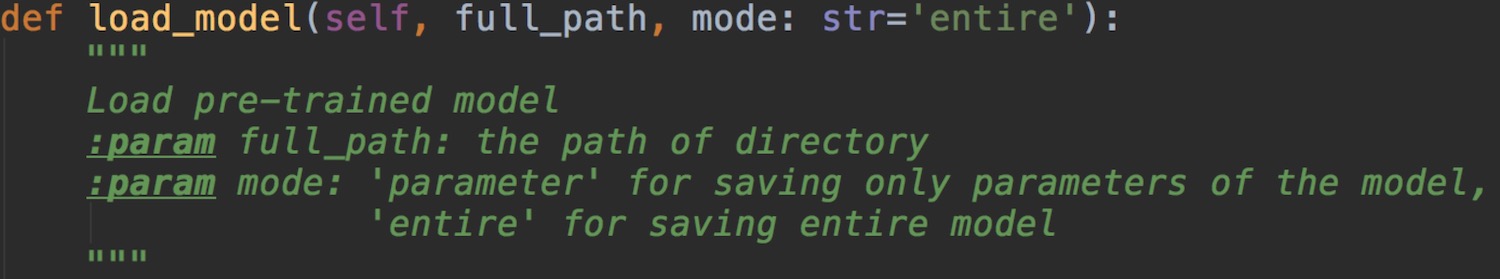}
    \caption{The description of \texttt{model\_load}.}\label{fig:load}
    \end{figure}
    \item \texttt{print\_info:} print basic information of model
    \item \texttt{plot\_exogenous:} print exogenous intensity.
\end{itemize}
In PoPPy, the instance of this class implements an inhomogeneous Poisson process, in which the exogenous intensity is used as the intensity function.

\begin{figure}[h]
    \centering
    \subfigure[exogenous intensity]{
    \includegraphics[width=0.4\textwidth]{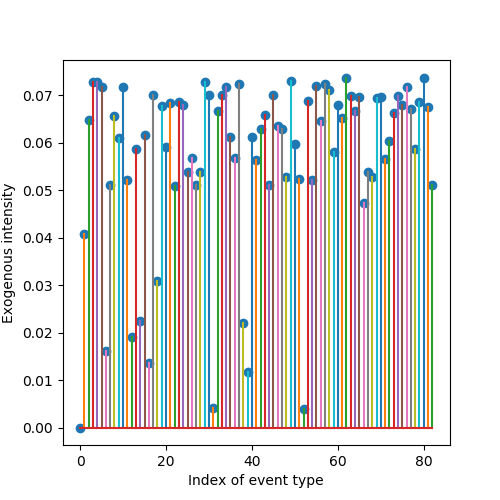}
    }
    \subfigure[Granger causality]{
    \includegraphics[width=0.4\textwidth]{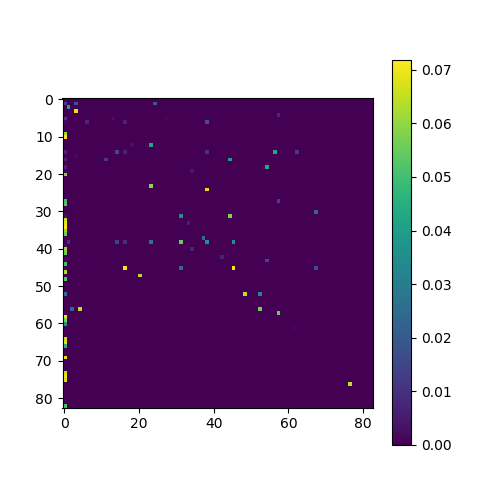}
    }
    \caption{Typical visualization results.}\label{fig:visual}
\end{figure}

An important subclass of this class is \texttt{model.HawkesProcess.HawkesProcessModel}. 
This subclass inherits most of the functions above except \texttt{print\_info} and \texttt{plot\_exogenous}. 
Additionally, because the Hawkes process considers the triggering patterns among different event types, this subclass has a new function \texttt{plot\_causality}, which plots the adjacency matrix of the event types' Granger causality graph. 
The typical visualization results of the exogenous intensity of different event types and the Granger causality among them are shown in Fig.~\ref{fig:visual}.

Compared with its parant class, \texttt{model.HawkesProcess.HawkesProcessModel} uses a specific intensity function, which is defined in the class  \texttt{model.HawkesProcess.HawkesProcessIntensity}.

\subsection{\texttt{model.HawkesProcess.HawkesProcessIntensity}}
This class inherits the functions in \texttt{torch.nn.Module}.
It defines the intensity function of a generalized Hawkes process, which contains the following functions:
\begin{itemize}
    \item \texttt{print\_info:} print the basic information of the intensity function.
    \item \texttt{intensity:} calculate $\lambda_{c_i}(t_i)$ of the $i$-th sample in the batch sampled by \texttt{EventSampler}.
    \item \texttt{expected\_counts:} calculate $\int_{t_{i-1}}^{t_i}\lambda_{c}(s)ds$ for $c\in\mathcal{C}$ and for the $i$-th sample in the batch.
    \item \texttt{forward:} override the forward function in \texttt{torch.nn.Module}. It calculates $\lambda_{c_i}(t_i)$ and  $\int_{t_{i-1}}^{t_i}\lambda_{c}(s)ds$ for $c\in\mathcal{C}$ for SGD. 
\end{itemize}

Specifically, the intensity function of type-$c$ event at time $t$ is defined as
\begin{eqnarray}
\begin{aligned}
\lambda_c(t)& = g_{\lambda}\left(\underbrace{\mu_c(\bm{f}_c, \bm{f}_s)}_{\mbox{exogenous intensity}} + \underbrace{\sum_{t_i<t}\phi_{cc_i}(t-t_i, \bm{f}_c, \bm{f}_{c_i})}_{\mbox{endogeneous impact}}\right)\\ 
&= \mu_c(\bm{f}_c, \bm{f}_s) + \sum_{t_i<t}\sum_{m=1}^{M}\alpha_{cc_i m}(\bm{f}_c, \bm{f}_{c_i})\kappa_m (t - t_i).
\end{aligned}
\end{eqnarray}
Here, the intensity function is consist of two parts:
\begin{itemize}
    \item \textbf{Exogenous intensity} $\mu_c(\bm{f}_c, \bm{f}_s)$: it is independent with time, which measures the intensity contributed by the intrinsic properties of sequence and event type. 
    \item \textbf{Endogenous impact} $\sum_{t_i<t}\phi_{cc_i}(t-t_i, \bm{f}_c, \bm{f}_{c_i})$: it sums up the influences of historical events quantitatively via \textbf{impact functions} $\{\phi_{cc'}(t)\}_{c,c'\in\mathcal{C}}$, which measures the intensity contributed by the historical observations. 
\end{itemize}
Furthermore, the impact function is decomposed with the help of basis representation, where $\kappa_m(t)$ is called the $m$-th \textbf{decay kernel} and $\alpha_{cc_i m}(\bm{f}_c, \bm{f}_{c_i})$ is the corresponding \textbf{coefficient}.

$g_{\lambda}(\cdot)$ is an activation function, which can be 
\begin{itemize}
    \item \textbf{Identity:} $g(x)=x$.
    \item \textbf{ReLU:} $g(x)=\max\{x, 0\}$.
    \item \textbf{Softplus:} $g(x)=\frac{1}{\beta}\log(1+\exp(-\beta x))$.
\end{itemize}

PoPPy provides multiple choices to implement various intensity functions --- each module can be parametrized in different ways. 

\subsubsection{\texttt{model.ExogenousIntensity.BasicExogenousIntensity}}
This class and its subclasses in \texttt{model.ExogenousIntensityFamily} implements several models of exogenous intensity, as shown in Table~\ref{tab:exo}.
\begin{table}[h]
\caption{Typical models of exogenous intensity.}\label{tab:exo}
\footnotesize{
    \centering
    \begin{tabular}{
        @{\hspace{2pt}}l@{\hspace{2pt}}|
        @{\hspace{2pt}}l@{\hspace{2pt}}
        }
        \hline\hline
        Class &Formulation \\ \hline
        \texttt{ExogenousIntensity.BasicExogenousIntensity} & $\mu_c(\bm{f}_c, \bm{f}_s)=\mu_c$\\
        \texttt{ExogenousIntensityFamily.NaiveExogenousIntensity} & $\mu_c(\bm{f}_c, \bm{f}_s)=g(\mu_c)$\\
        \texttt{ExogenousIntensityFamily.LinearExogenousIntensity} & $\mu_c(\bm{f}_c, \bm{f}_s)=g(\bm{w}_c^{\top}\bm{f}_s)$\\
        \texttt{ExogenousIntensityFamily.NeuralExogenousIntensity} & $\mu_c(\bm{f}_c, \bm{f}_s)=NN(\bm{f}_c, \bm{f}_s)$\\
        \hline\hline
      \end{tabular}
}
\end{table}

Here, the activation function $g(\cdot)$ is defined as aforementioned $g_{\lambda}$. 

Note that the last two models require event and sequence features as input. 
When they are called while the features are not given, PoPPy will add one more embedding layer to generate event/sequence features from their index, and learn this layer during training.

\subsubsection{\texttt{model.EndogenousImpact.BasicEndogenousImpact}}
This class and its subclasses in \texttt{model.EndogenousImpactFamily} implement several models of the coefficients of the impact function, as shown in Table~\ref{tab:end}.
\begin{table}[h]
\caption{Typical models of endogenous impact's coefficient.}\label{tab:end}
\footnotesize{
    \centering
    \begin{tabular}{
        @{\hspace{2pt}}l@{\hspace{2pt}}|
        @{\hspace{2pt}}l@{\hspace{2pt}}
        }
        \hline\hline
        Class &Formulation \\ \hline
        \texttt{EndogenousImpact.BasicEndogenousImpact} & $\alpha_{cc'm}(\bm{f}_c, \bm{f}_{c'})=\alpha_{cc'm}$\\
        \texttt{EndogenousImpactFamily.NaiveEndogenousImpact} & $\alpha_{cc'm}(\bm{f}_c, \bm{f}_{c'})=g(\alpha_{cc'm})$\\
        \texttt{EndogenousImpactFamily.FactorizedEndogenousImpact} & $\alpha_{cc'm}(\bm{f}_c, \bm{f}_{c'})=g(\bm{u}_{cm}^{\top}\bm{v}_{c'm})$\\
        \texttt{EndogenousImpactFamily.LinearEndogenousImpact} & $\alpha_{cc'm}(\bm{f}_c, \bm{f}_{c'})=g(\bm{w}_{cm}^{\top}\bm{f}_{c'})$\\
        \texttt{EndogenousImpactFamily.BiLinearEndogenousImpact} & $\alpha_{cc'm}(\bm{f}_c, \bm{f}_{c'})=g(\bm{f}_{c}^{\top}\bm{W}_{m}\bm{f}_{c'})$ \\
        \hline\hline
      \end{tabular}
}
\end{table}

Here, the activation function $g(\cdot)$ is defined as aforementioned $g_{\lambda}$. 

Note that the last two models require event and sequence features as input. 
When they are called while the features are not given, PoPPy will add one more embedding layer to generate event/sequence features from their index, and learn this layer during training.

\subsubsection{\texttt{model.DecayKernel.BasicDecayKernel}}
This class and its subclasses in \texttt{model.DecayKernelFamily} implements several models of the decay kernel, as shown in Table~\ref{tab:ker}.
\begin{table}[h]
\caption{Typical models of decay kernel.}\label{tab:ker}
\footnotesize{
    \centering
    \begin{tabular}{
        @{\hspace{2pt}}l@{\hspace{2pt}}|
        @{\hspace{2pt}}l@{\hspace{2pt}}|
        @{\hspace{2pt}}l@{\hspace{2pt}}
        }
        \hline\hline
        Class &$M$ &Formulation \\ \hline
        \texttt{DecayKernelFamily.ExponentialKernel}~\cite{zhou2013learning} &1 
        & $\kappa(t) =\begin{cases}
        \omega\exp(-\omega(t-\delta)),& t\geq\delta,\\
        0,& t<\delta\end{cases}$\\
        \texttt{DecayKernelFamily.RayleighKernel} &1 
        & $\kappa(t)=\omega t\exp(-\omega t^2/s)$\\
        \texttt{DecayKernelFamily.GaussianKernel} &1 & $\kappa(t)=\frac{1}{\sqrt{2\pi}\sigma} \exp(-\frac{t^2}{2\sigma^2})$\\
        \texttt{DecayKernelFamily.PowerlawKernel}~\cite{zhao2015seismic} &1 
        & $\kappa(t) =\begin{cases}
        (\omega-1)\delta^{\omega-1}t^{-\omega},& x\geq\delta,\\
        (\omega-1)/\delta,& 0<x<\delta\end{cases}$\\
        \texttt{DecayKernelFamily.GateKernel} &1 & $\kappa(t)=\frac{1}{\delta},~t\in [\omega, \omega+\delta]$ \\
        \texttt{DecayKernelFamily.MultiGaussKernel}~\cite{xu2016learning} &>1 & $\kappa_m(t) = \frac{1}{\sqrt{2\pi}\sigma_m} \exp(-\frac{(t-t_m)^2}{2\sigma_m^2})$ \\
        \hline\hline
      \end{tabular}
}
\end{table}

\begin{figure}[h]
\centering
\subfigure[Exponential kernel]{
\includegraphics[width=0.3\textwidth]{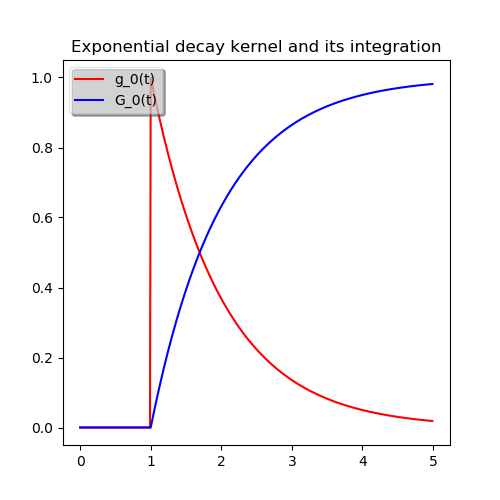}
}
\subfigure[Rayleigh kernel]{
\includegraphics[width=0.3\textwidth]{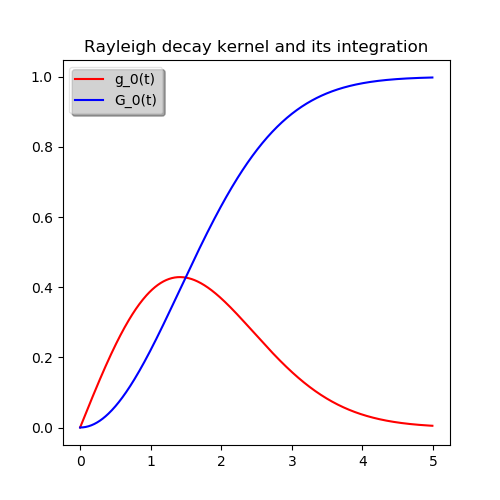}
}
\subfigure[Gaussian kernel]{
\includegraphics[width=0.3\textwidth]{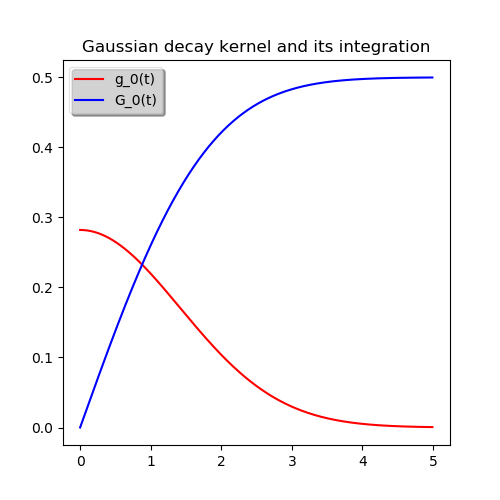}
}
\subfigure[Powerlaw kernel]{
\includegraphics[width=0.3\textwidth]{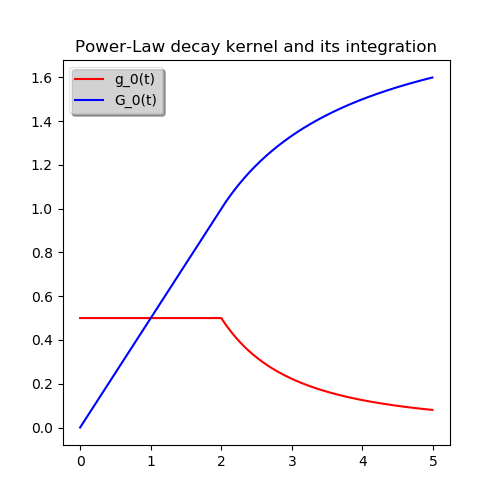}
}
\subfigure[Gate kernel]{
\includegraphics[width=0.3\textwidth]{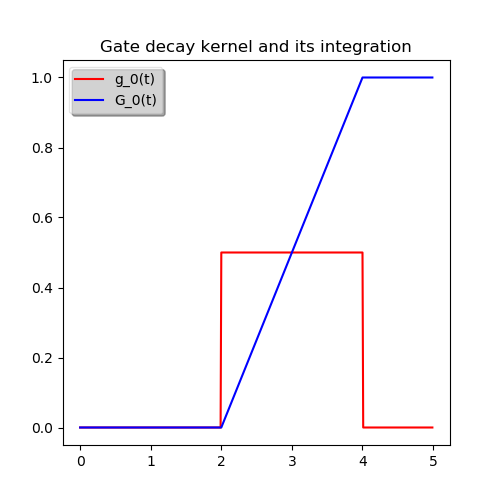}
}
\subfigure[Multi-Gaussian kernel]{
\includegraphics[width=0.3\textwidth]{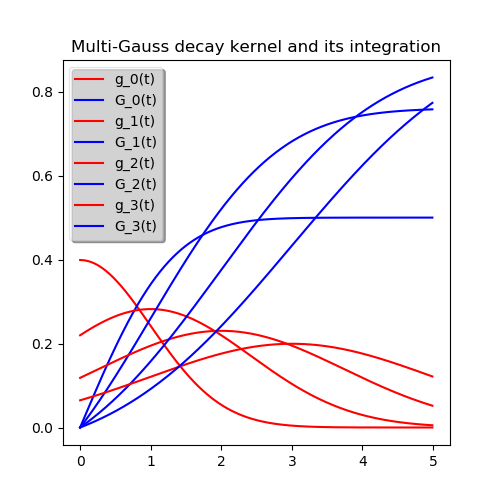}
}
\caption{Examples of decay kernels and their integration values.}\label{fig:kernel}
\end{figure}

Fig.~\ref{fig:kernel} visualizes some examples.

\section{Learning Algorithm}
\subsection{Loss functions}
With the help of PyTorch, PoPPy learns the point process models above efficiently by stochastic gradient descent on CPU or GPU~\cite{mei2017neural}.
Different from existing point process toolboxes, which mainly focuses on the maximum likelihood estimation of point process models, PoPPy integrates three loss functions to learn the models, as shown in Table~\ref{tab:loss}.
\begin{table}[h]
\caption{A list of loss functions.}\label{tab:loss}
\centering{
\footnotesize{
    \centering
    \begin{tabular}{
        @{\hspace{2pt}}l@{\hspace{2pt}}
        }
        \hline\hline
        Maximum Likelihood Estimation~\cite{zhou2013learning,xu2016learning}\\ 
        - Class: \texttt{OtherLayers.MaxLogLike} \\
        - Formulation: $L(\theta) = -\sum_{i\in \mathcal{D}}\left(\log\lambda_{c_i}(t_i)-\sum_{c\in\mathcal{C}}\int_{t_{i-1}}^{t_i}\lambda_c(s)ds\right)$\\ \hline
        Least Square Estimation~\cite{xu2018benefits,xu2018online}\\
        - Class: \texttt{OtherLayers.LeastSquare}\\
        - Formulation: $L(\theta) = \sum_{i\in \mathcal{D}}\|\int_{t_{i-1}}^{t_i}\bm{\lambda}(s)ds - \bm{1}_{c_i}\|_2^2$\\ \hline
        Conditional Likelihood Estimation~\cite{xu2017patient}\\
        - Class: \texttt{OtherLayers.CrossEntropy}\\
        - Formulation: $L(\theta) = -\sum_{i\in \mathcal{D}}\log p(c_i | t_i, \mathcal{H}_i) = -\sum_{i\in \mathcal{D}}\log softmax\left(\int_{t_{i-1}}^{t_i}\bm{\lambda}(s)ds\right).$\\
        \hline\hline
      \end{tabular}
}
}
\end{table}

Here $\bm{\lambda}(t)=[\lambda_1(t), ..., \lambda_{|\mathcal{C}|}(t)]$ and $\bm{1}_c$ is an one-hot vector whose the $c$-th element is 1.

\subsection{Stochastic gradient decent}
All the optimizers and the learning rate schedulers in PyTorch are applicable to PoPPy. 
A typical configuration is using Adam + Exponential learning rate decay strategy, which should achieve good learning results in most situations. 
The details can be found in the demo scripts in the folder \texttt{example}.

\textbf{Trick:} Although most of the optimizers are applicable, generally, Adam achieves the best performance in our experiments~\cite{mei2017neural}. 

\subsection{Optional regularization}
Besides the L2-norm regularizer in PyTorch, PoPPy provides two more regularizers when learning models.
\begin{enumerate}
    \item \textbf{Sparsity:} L1-norm of model's parameters can be applied to the models, which helps to learn structural parameters.
    \item \textbf{Nonnegativeness:} If it is required, PoPPy can ensure the parameters to be nonnegative during training.
\end{enumerate}

\textbf{Trick:} When the activation function of impact coefficient is softplus, you'd better close the nonnegative constraint by setting the input \texttt{nonnegative} of the function \texttt{fit} as \textbf{None}.

\section{Examples}
\begin{figure}[h]
\centering
\includegraphics[width=0.7\textwidth]{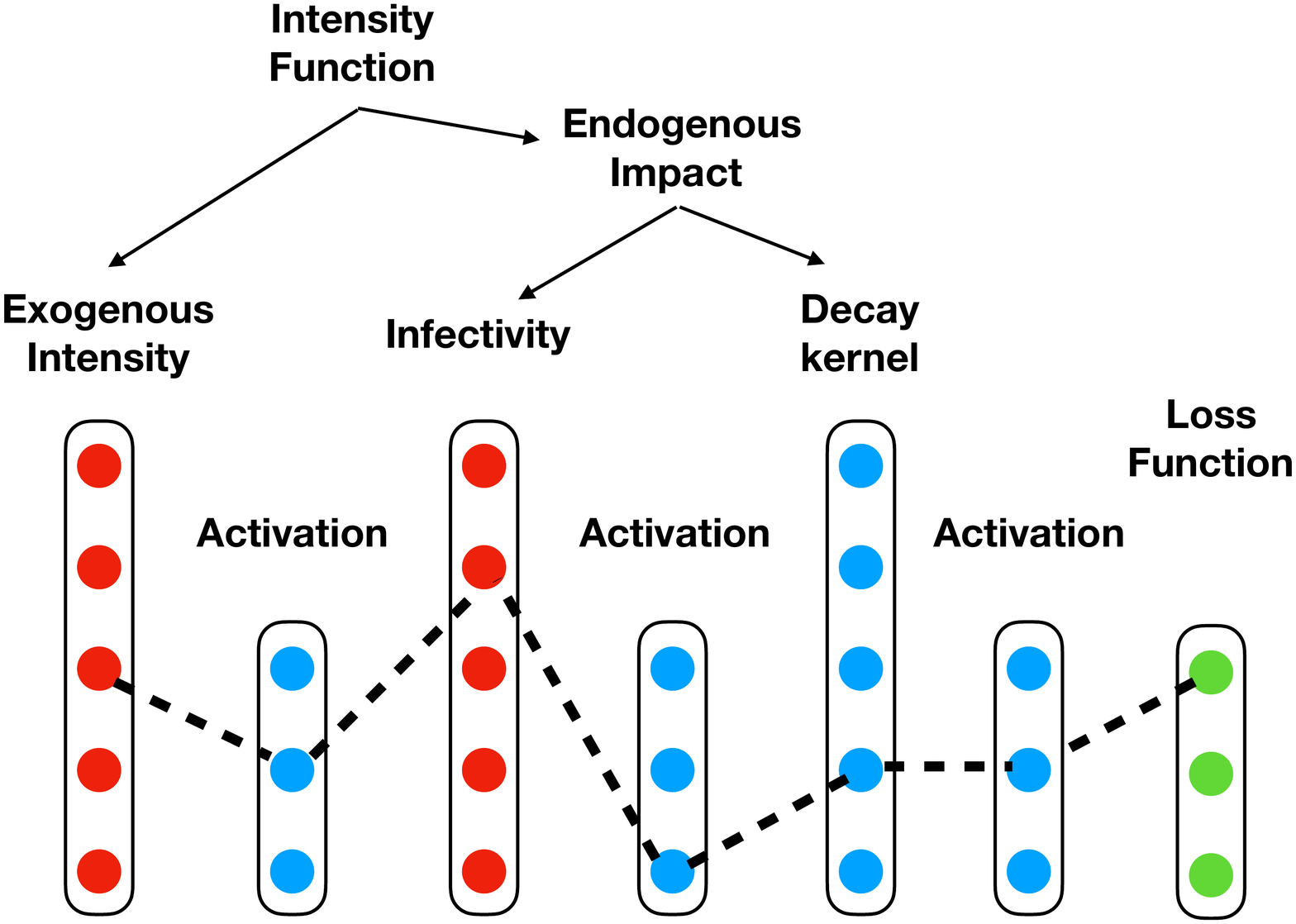}
\caption{Illustration the contruction of a point process model.}\label{fig:model}
\end{figure}

As a result, using PoPPy, users can build their own point process models by combining different modules with high flexibility. 
As shown in Fig.~\ref{fig:model}, Each point process model can be built by selecting different modules and combining them. The red dots represent the module with learnable parameters, the blue dots represent the module without parameters, and the green dots represent loss function modules. 
Moreover, users can add their own modules and design specific point process models for their applications quickly, as long as the new classes override the corresponding functions.

Finally, we list some typical models implemented by PoPPy in Table~\ref{tab:example}.
\begin{table}[h]
\caption{Typical models implemented by PoPPy.}\label{tab:example}
\centering{
\footnotesize{
    \centering
    \begin{tabular}{
        @{\hspace{2pt}}l@{\hspace{2pt}}|@{\hspace{2pt}}l@{\hspace{2pt}}
        }
        \hline\hline
        Model & Linear Hawkes process~\cite{zhou2013learning}\\
        Exogenous Intensity      &\texttt{NaiveExogenousIntensity}\\
        Endogenous Impact        &\texttt{NavieEndogenousImpact}\\
        Decay Kernel             &\texttt{ExponentialKernel}\\
        Activation $g_{\lambda}$ &Identity\\
        Loss                     &\texttt{MaxLogLike}\\ \hline
        Model & Linear Hawkes process~\cite{xu2016learning,xu2018superposition}\\
        Exogenous Intensity      &\texttt{NaiveExogenousIntensity}\\
        Endogenous Impact        &\texttt{NavieEndogenousImpact}\\
        Decay Kernel             &\texttt{MultiGaussKernel}\\
        Activation $g_{\lambda}$ &Identity\\
        Loss                     &\texttt{MaxLogLike}\\ \hline
        Model & Linear Hawkes process~\cite{xu2018benefits}\\
        Exogenous Intensity      &\texttt{NaiveExogenousIntensity}\\
        Endogenous Impact        &\texttt{NavieEndogenousImpact}\\
        Decay Kernel             &\texttt{MultiGaussKernel}\\
        Activation $g_{\lambda}$ &Identity\\
        Loss                     &\texttt{LeastSquares}\\ \hline
        Model & Factorized point process~\cite{xu2018online}\\
        Exogenous Intensity      &\texttt{LinearExogenousIntensity}\\
        Endogenous Impact        &\texttt{FactorizedEndogenousImpact}\\
        Decay Kernel             &\texttt{ExponentialKernel}\\
        Activation $g_{\lambda}$ &Identity\\
        Loss                     &\texttt{LeastSquares}\\ \hline
        Model & Semi-Parametric Hawkes process~\cite{engelhard2018predicting}\\
        Exogenous Intensity      &\texttt{LinearExogenousIntensity}\\
        Endogenous Impact        &\texttt{NavieEndogenousImpact}\\
        Decay Kernel             &\texttt{MultiGaussKernel}\\
        Activation $g_{\lambda}$ &Identity\\
        Loss                     &\texttt{MaxLogLike}\\ \hline
        Model & Parametric self-correcting process~\cite{xu2015trailer}\\
        Exogenous Intensity      &\texttt{LinearExogenousIntensity}\\
        Endogenous Impact        &\texttt{LinearEndogenousImpact}\\
        Decay Kernel             &\texttt{GateKernel}\\
        Activation $g_{\lambda}$ &Softplus\\
        Loss                     &\texttt{MaxLogLike}\\ \hline
        Model & Mutually-correcting process~\cite{xu2017patient}\\
        Exogenous Intensity      &\texttt{LinearExogenousIntensity}\\
        Endogenous Impact        &\texttt{LinearEndogenousImpact}\\
        Decay Kernel             &\texttt{GaussianKernel}\\
        Activation $g_{\lambda}$ &Softplus\\
        Loss                     &\texttt{CrossEntropy}\\ \hline
        \hline\hline
      \end{tabular}
}
}
\end{table}

\bibliographystyle{ieee}
\bibliography{example_paper}

\end{document}